%% file: iclr2026_conference__1_.tex
\documentclass{article} % For LaTeX2e
\usepackage{iclr2026_conference,times}

\input{math_commands.tex}

\usepackage{hyperref}
\usepackage{url}

\usepackage{graphicx}
\usepackage{booktabs}
\usepackage{multirow}
\usepackage{xcolor}
\usepackage{wrapfig}
\usepackage[most]{tcolorbox}
\usepackage{amssymb}
\usepackage{amsthm}

\newtheorem{theorem}{Theorem}

\title{Short Horizons and Sparse Concepts: a Mathematical View of the Readout in the J-lens}

\author{
\centerline{\textbf{Shi-Qi Yan$^{*}$,\qquad Kai-Xuan Ding$^{*}$,\qquad Chao-Hong Tan,\qquad Qian Chen,}} \\
\centerline{\textbf{Wen Wang,\qquad Xiangang Li,\qquad Zhen-Hua Ling}} \\
Alibaba Token Hub, Alibaba Group\\
$^*$Equal Contribution\\
\texttt{sqyan01@mail.ustc.edu.cn}
}

\iclrfinalcopy % Uncomment for camera-ready version, but NOT for submission.
\begin{document}

\maketitle

\input{0_Abstract}
\input{1_Introduction}
\input{2_Related_Work}
\input{3_Preliminaries}

\input{4_Methods}
\input{5_Experiments}
\input{6_Conclusion}

\bibliography{iclr2026_conference}
\bibliographystyle{iclr2026_conference}

\end{document}

%% file: math_commands.tex
\usepackage{amsmath,amsfonts,bm}

\def\eqref#1{equation~\ref{#1}}
\def\1{\bm{1}}

\DeclareMathAlphabet{\mathsfit}{\encodingdefault}{\sfdefault}{m}{sl}
\SetMathAlphabet{\mathsfit}{bold}{\encodingdefault}{\sfdefault}{bx}{n}

%% file: 0_Abstract.tex
\begin{abstract}
The Jacobian lens (J-lens) has been proposed as a way to read verbalizable representations from language models. However, its principle and meaning lack a detailed and theoretical discussion. We provide a mathematical view of this interpretation and of its assumed causal structure. Besides treating the J-lens as a heuristic probe, we further regard it as a first-order causal transfer operator from intermediate activations to expected future readouts. We study the Jacobian matrix as the optimal local linear approximation of the downstream mapping, analyze its global approximation behavior and bias, and identify its mathematical meaning as an expectation over anticipated future readouts. Further analysis of the Jacobian energy distribution reveals that its causal geometry is highly sparse. The energy decays with depth, concentrates in an extremely small proportion, and decomposes into diagonal pathways and specific critical positions. This decomposition further resolves the expectation of the J-lens over future outputs into short-horizon and sparse concept predictions, providing a more intuitive attribution and explanation for the ability of the J-lens to visualize concepts during the thinking process. Based on the theory, we propose a simple but effective improvement strategy and decoupling method for the J-lens, which significantly enhances the ability of the J-lens to read out correct intermediate concepts.
\end{abstract}

%% file: 1_Introduction.tex
\section{Introduction}\label{sec:intro}

Most computation in large language models (LLMs) occurs in the background~\citep{fewshot,llmemergent}. A prompt may elicit a brief answer, but this answer arises from a sequence of complex intermediate transformations whose content cannot be directly observed. Mechanistic interpretability attempts to open this black box by reading internal states, intervening on them, and relating them to behavior~\citep{openproblem,2016probes,2019probes}. A central difficulty is that intermediate-layer representations are not naturally expressible in the vocabulary of human concepts, and their connection to the output layer is indirect. The residual stream is a high-dimensional vector space, whereas interpretation requires a structure that is sparse, reportable, and causally meaningful.

The recently proposed Jacobian lens, or J-lens, addresses this difficulty by propagating a hidden state along the downstream computation of the model and measuring how its perturbation affects the final-layer state~\citep{jlens}. 
The resulting J-space is described as sparse and verbalizable: hidden states can be approximated by nonnegative combinations of a few J-lens vectors that support reporting, modulation, and selective computation~\citep{olsson2022induction}. 
However, it lacks a clear and detailed mathematical interpretation. The reasons why averaging Jacobian matrices over future positions yields valid tokens, and how the resulting representations relate to the internal thoughts of the model remain unknown. The J-lens claims that the first-order sensitivity of future outputs to hidden states is itself a carrier of verbalizable tendencies. Evaluating this claim requires establishing in what sense the Jacobian matrix is optimal, what objective it approximates, and whether the causal influence behind the average is uniformly distributed or concentrated at specific positions.

To address this issue, we provide a clear mathematical interpretation of the J-lens and empirically validate its effectiveness and identify its practical deviations. Given an input $h_{t,l}$, the J-lens can be viewed as a global fit of the first-order linear approximation to the latent mapping $\mathbb{E}_{t' > t}[h_{t',L}]$. Conceptually, we regard the J-lens as an averaged first-order causal transfer operator. In a given context, the Jacobian matrix is the optimal local linear approximation of how a small change in an intermediate activation affects subsequent outputs. Under this formulation, the residual vanishes to first order. Averaging this operator over samples yields the practical $J_\ell$, which provides the globally optimal approximation under the minimum mean squared error criterion. We use a Stein-based identity to justify the passage from local approximation to global fitting, estimate the potential theoretical bias, and provide experimental evidence that corroborates this conclusion from the reverse direction. 

Building on this, we further provide an attribution of the causal mapping fitted by the J-lens. The averaged operator $J_\ell$ compresses all position-to-position Jacobian matrices into a single matrix, but the underlying sensitivity is not uniform. We therefore study the Jacobian energy $E_L(t, t')$ between a source position $t$ and a target position $t'$. Experiments validate the sparsity of the Jacobian energy: on the one hand, the energy tends to be larger in earlier layers and much smaller near the output; on the other hand, within a single layer, the Jacobian energy exhibits strong concentration, and a small number of pairs carry most of the total energy. Through visualization, we observe that the concentration of energy can be grouped into two patterns: the diagonal pattern ($t \approx t'$) and the horizontal or vertical pattern ($t=t^*$ or $t'=t^*$). They represent two effective outcomes of the J-lens mapping in practice: short-horizon token prediction and highly sensitive tokens at critical positions. We refer to these two outcomes as the short horizons and sparse concepts. The readout results of the practical J-lens also effectively corroborate our conclusion. 

Based on the above theoretical and experimental results, we design experiments to further validate the effectiveness of our interpretation by improving the J-lens. First, the concentration effect of the Jacobian energy indicates that the predictions of the J-lens naturally align with short-horizon tokens and sparse concepts. Averaging the Jacobian matrices over all positions turns unimportant positions into noise and affects the practical representations of the J-lens. We therefore introduce a simple but effective filtering strategy that uses only a subset of positions with the highest energy in the J-lens construction. Experiments show that using only the Jacobian matrices from the top 10\% to 20\% of positions by energy significantly improves the capability of the J-lens. Second, we also attempt to decouple the two representations of the J-lens~\citep{ghandeharioun2024patchscopes,conmy2023acdc}. By designing masks targeting the diagonal~\citep{geiger2021causal,meng2022locating}, the two capabilities are decoupled by removing or retaining only the Jacobian matrices on diagonal entries. However, experiments show that the coupling between the two capabilities is stronger than expected, and the strengthening or weakening of the two capabilities exhibits a positive correlation. As this work is still in progress, this becomes a goal of our future research.

In summary, this work makes three contributions. First, we provide a mathematical foundation for the J-lens and establish it as the expectation over anticipated outputs. Through local linearization, global least-squares projection, and a Stein-type bridge, we connect the J-lens to the optimal linear readout and explicitly characterize non-Gaussian and nonlinear deviations. Second, we present experimental analyses and visualization studies that cluster the expectation over anticipated outputs into short-horizon token predictions and critical concept predictions. Third, we design a lens improving and decoupling method that further validates the reliability and effectiveness of our conclusions.

%% file: 2_Related_Work.tex
\section{Background: J-lens and Global Workspace}\label{sec:background}

A transformer language model maintains, at each token position, a residual-stream vector that is read from and written to by every layer. Early states are close to the input token, whereas final states are directly mapped to vocabulary scores by the unembedding matrix $W_U$. Intermediate states are therefore computational, but not directly verbal. The logit lens reads them as if all layers shared one coordinate system; this works near the output but often fails earlier, where representations have not yet been rotated into output coordinates~\citep{logitlens,belrose2023tunedlens,geva2022ffn}.

The Jacobian lens (J-lens) instead reads an intermediate activation by its average first-order effect on present and future outputs~\citep{jlens}. For layer $\ell$, it averages the Jacobian from a source activation to later final-layer states over source positions, target positions, and prompts:
\begin{equation}
J_\ell=\mathbb{E}_{t,t'\ge t,\mathrm{prompt}}
\left[\frac{\partial h_{\mathrm{final},t'}}{\partial h_{\ell,t}}\right],
\qquad
\mathrm{lens}(h_\ell)=\mathrm{softmax}\!\left(W_U\,\mathrm{norm}(J_\ell h_\ell)\right).
\label{eq:background-jlens}
\end{equation}
The top tokens of this readout describe what an activation is disposed to say across contexts, rather than what one context happens to say. The rows of $W_UJ_\ell$ define token-associated readout directions; their sparse nonnegative combinations form the J-space.

Functionally, it supports five properties corresponding to conscious access: verbal report, directed modulation, internal reasoning, flexible generalization, and selectivity. However, a more detailed theoretical account is missing, and the J-lens itself is approximate and limited in applications. This paper takes this challenge seriously. We investigate why the averaged Jacobian matrix can approximate the optimal readout, how the approximation is affected by nonlinear and non-Gaussian bias, and what sparse causal structure over positions underlies this interpretation.

%% file: 3_Preliminaries.tex
\section{Mathematical Explanation of J-lens: Expectation of the Future Readouts}\label{sec:prelim}

% \subsection{Object of study}
Given a layer $\ell$ and position $t$. We study the relationship between the J-lens and the mapping function of the averaged future readouts:
\begin{equation}
y_t=f_\ell(h_{\ell,t}) := \mathbb{E}_{t'\ge t}[h_{\mathrm{final},t'}],
\qquad
f_\ell:\mathbb{R}^d\to\mathbb{R}^d,
\end{equation}
where $h_{\ell,t}$ is the hidden states at layer $\ell$ and position $t$, and $y_t$ is an averaged future final-layer responses $h_{\mathrm{final},t'}|_{t'\ge t}$. 
The function $f_\ell$ is implicitly defined by the remaining network and is generally nonlinear.
In this section, we systematically analyze the mathematical principles underlying the J-lens, corroborate its first-order approximation to the nonlinear mapping $f_\ell$, and confirm its meaning. 
Specifically, given the local function-fitting capability of the Jacobian matrix, we mathematically interpret the construction of the J-lens, i.e., the statistical expectation of the Jacobian over the training distribution, as a process of fitting $f_\ell$ from local approximation to global~\citep{stein1981estimation}. 
In addition, we analyze its approximation bias from both theoretical and experimental perspectives, validating the reliability and effectiveness of the theoretical interpretation in this section while providing a new perspective on the failure of the J-lens in early layers.

% The function $f_\ell$ is implicitly defined by the remaining network and is generally nonlinear. A first-order theory replaces $f_\ell$ by an affine map, but there are two distinct notions of replacement requiring further discussion.

\subsection{Local View: Tangent Linearization}
For a smooth map $f:\mathbb{R}^d\to\mathbb{R}^d$, the Jacobian $J_f(h_0)\in\mathbb{R}^{d\times d}$ has entries $[J_f(h_0)]_{ij}=\partial f_i/\partial h_j$ at $h_0$. The first-order Taylor expansion is
\begin{equation}
f(h_0+\delta)=f(h_0)+J_f(h_0)\cdot\delta+o(\|\delta\|).
\end{equation}
Thus the Jacobian can be regarded as the local sensitivity map: a small intervention $\delta$ at $h_{\ell,t}$ changes the predicted future response, to first order, by $J_{f_\ell}(h_{\ell,t}) \cdot \delta$. The approximation is asymptotically exact as $\|\delta\|\to0$, but its usable radius shrinks where curvature is large, and the matrix is pointwise rather than global. Therefore, a single Jacobian matrix solely estimates the first-order approximation with tangent linearization in a local range.

\subsection{Global View: Least-squares Readout}
The global view fixes one affine map for the whole distribution. Let $h\sim p$, $y=f(h)$, $\mu=\mathbb{E}[h]$, and $\Sigma=\operatorname{Cov}(h)$, with $\Sigma\succ0$. Writing $\tilde h=h-\mu$ and $\tilde y=y-\mathbb{E}[y]$, a global first-order approximation of the function $f$ can be treated as a least-squares estimation:
\begin{equation}
\min_{W,b}\ \mathbb{E}\|y-(Wh+b)\|^2
\end{equation}
with the normal-equation solution
\begin{equation}
\begin{cases}
W^*=\mathbb{E}[\tilde y\,\tilde h^\top]\,\Sigma^{-1}=\operatorname{Cov}(y,h)\operatorname{Cov}(h)^{-1}\\
b^*=\mathbb{E}[y]-W^*\mu.
\end{cases}
\end{equation}
The residual is orthogonal to all affine functions of $h$, so $W^*h+b^*$ is the $L^2$ projection of the regression function $m(h)=\mathbb{E}[y\mid h]$ onto affine functions. This optimum requires no Gaussianity and no differentiability; it is a statistical compromise slope over the data distribution.

\subsection{Score identity and the Stein bridge}\label{subsec:stein}

The link between local derivatives and global least squares is integration by parts. Assume that $h$ has a smooth positive density $p$ on $\mathbb{R}^d$, that $f:\mathbb{R}^d\to\mathbb{R}^d$ is componentwise differentiable, and that $f$ grows slowly enough relative to $p$ that boundary terms vanish~\citep{hyvarinen2005score}. The score function is
\begin{equation}
\label{eq:score-function}
s(h):=\nabla\log p(h)=\frac{\nabla p(h)}{p(h)} .
\end{equation}
For one component $f_i$ and one input coordinate $h_j$,
\begin{equation}
\mathbb{E}[\partial_j f_i(h)]=\int_{\mathbb{R}^d} \partial_j f_i(h)\,p(h)\,dh=-\int_{\mathbb{R}^d} f_i(h)\,\partial_j p(h)\,dh ,
\end{equation}
where the boundary term vanishes under the growth assumption, where $f_i(h)p(h)\to0$ as $\|h\|\to\infty$ for every component $i$. Since $\partial_j p=p\,\partial_j\log p=p\,s_j$ as shown in Eq.~\ref{eq:score-function},
\begin{equation}
\mathbb{E}[\partial_j f_i(h)]=-\mathbb{E}[f_i(h)s_j(h)] .    
\end{equation}
Stacking over $i$ and $j$ gives the multivariate score identity:
\begin{equation}
\mathbb{E}[J_f(h)]=-\mathbb{E}\!\left[f(h)s(h)^\top\right].
\label{eq:score-identity}
\end{equation}
The left side averages local slopes. The right side says that this average is determined by correlations between function values and the local tilt of the density. Derivative information is therefore encoded in how the density changes, not only in how the function bends.

Equation~\eqref{eq:score-identity} is distribution-free once the regularity conditions hold. The Gaussian case is special because the score is affine. If $h\sim\mathcal N(\mu,\Sigma)$ with $\Sigma\succ0$, then
\begin{equation}
\log p(h) = \text{const}-\frac12(h-\mu)^\top\Sigma^{-1}(h-\mu),\qquad
s(h)=-\Sigma^{-1}(h-\mu).
\end{equation}
Substituting into~\eqref{eq:score-identity},
\begin{equation}
\mathbb{E}[J_f(h)]=-\mathbb{E}\!\left[f(h)\left(-\Sigma^{-1}(h-\mu)\right)^\top\right]=\mathbb{E}\!\left[f(h)(h-\mu)^\top\right]\Sigma^{-1}.
\end{equation}
Because $\mathbb{E}[s(h)]=0$, replacing $f$ by $\tilde f=f-\mathbb{E}[f(h)]$ does not change the right-hand side. Hence
\begin{equation}
\mathbb{E}[J_f(h)]=\mathbb{E}[\tilde f(h)\tilde h^\top]\Sigma^{-1}
=\operatorname{Cov}(f(h),h)\Sigma^{-1}=W^* .
\end{equation}
This is the Stein bridge: under Gaussian inputs, the average Jacobian is exactly the population least-squares slope.

\begin{theorem}[Stein bridge~\citep{stein1981estimation}]\label{thm:stein-bridge}
Let $h\sim\mathcal N(\mu,\Sigma)$ with $\Sigma\succ0$, and let $f$ satisfy the mild growth and differentiability conditions above. Then
\begin{equation}
% \boxed{\;
\mathbb{E}[J_f(h)]=\operatorname{Cov}(f(h),h)\operatorname{Cov}(h)^{-1}=W^* .
% \;}
\end{equation}
\end{theorem}
Under Gaussian inputs, the averaged Jacobian matrix is exactly equal to the population least-squares slope. Therefore, the J-lens can be understood as an approximation to the optimal linear readout. When the input distribution is close to Gaussian, this approximation is accurate; when the input distribution deviates substantially from Gaussian, an additional residual term is introduced into the score function, inducing a systematic bias between the averaged Jacobian and the optimal linear readout. In the subsequent section, we will analyze this bias and evaluate the layer-level valid range of our theoretical interpretation.

\begin{figure}[t]
  \vspace{-5mm}
  \centering
  \includegraphics[width=1.0\textwidth]{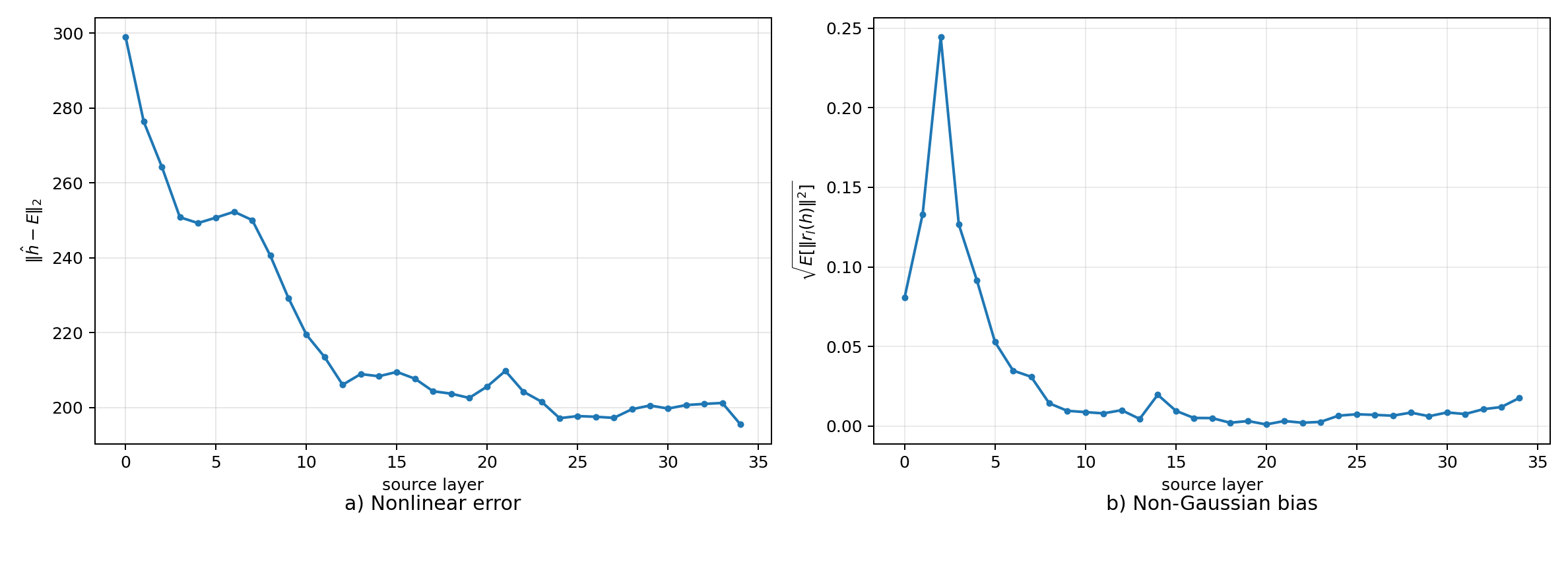}
  \caption{
  Layer-wise bias analysis of J-lens for Qwen3-8B.
  }
  \label{fig:bias-analysis}
\end{figure}

\subsection{Bias Analysis}
The J-lens involves two approximations. The first replaces the nonlinear mapping with a first-order linear transfer matrix, and the second approximates the globally optimal transfer matrix by the expectation of local Jacobians. 
Nonlinear and non-Gaussian biases are therefore systematically introduced and amplified as the distance between source and target layers increases. 
In this section, we analyze in detail the specific effects of these biases from both theoretical and experimental perspectives and provide a new explanation for the early failure of the J-lens.

\subsubsection{Nonlinear Bias}\label{subsec:nonlinear-bias}
The Stein bridge compares two slopes, $W^*$ and $\mathbb{E}[J_f(h)]$. A separate error remains even if the best possible slope is used: a linear map cannot reproduce curvature. Let $\mu=\mathbb{E}[h]$, $\delta=h-\mu$, $J=J_f(\mu)$, and let $H_i=\nabla^2 f_i(\mu)$ be the Hessian of output component $i$. Write $H[\delta,\delta]$ for the vector with components $\delta^\top H_i\delta$ and $H[\Sigma]=\mathbb{E}[H[\delta,\delta]]$, whose $i$-th component is $\operatorname{tr}(H_i\Sigma)$. The intrinsic nonlinear error of the optimal affine readout is
\begin{equation}
    \varepsilon_{\mathrm{lin}}^2=\mathbb{E}\big\| f(h)-\big(\mathbb{E}[f(h)]+W^*(h-\mu)\big)\big\|^2 .
\end{equation}
This is the residual that no choice of linear slope can remove.

\subsubsection{Non-Gaussian Bias}
Away from Gaussianity, decompose the score into its Gaussian part plus a residual:
\begin{equation}
    s(h)=-\Sigma^{-1}(h-\mu)+r(h),
\qquad
r(h)=0\ \Longleftrightarrow\ h\ \text{is Gaussian}.
\end{equation}
Substitution into the score identity gives the exact bias formula
\begin{equation}
\epsilon_\text{Gau}:=W^*-\mathbb{E}[J_f(h)]
=
\mathbb{E}\!\left[\tilde f(h)\,r(h)^\top\right].    
\end{equation}
The gap is therefore not controlled by nonlinearity alone. It is the covariance between the centered output fluctuation and the pointwise non-Gaussianity of the density. A standard Cauchy--Schwarz bound yields
\begin{equation}
\|\epsilon_\text{Gau}\|_F \le \sqrt{\mathbb{E}\|\tilde f(h)\|^2}\, \sqrt{\mathbb{E}\|r(h)\|^2}
\end{equation}
where the second factor is a distribution-level non-Gaussianity measure.

\subsubsection{Overall Bias}
As a cascade bias system without correlation, the overall system bias can be calculated by a sum of the squared biases:
\begin{equation}
    \epsilon_{\bar{J}}^2 = \epsilon_{\text{lin}}^2 + \|\epsilon_\text{Gau}\|_F^2.
\end{equation}
To verify the consistency between theory and experiments, we perform a layer-wise quantitative comparison between the theoretical error and the empirical error of the J-lens. The results show that both the nonlinear bias and the non-Gaussian error decrease as the source layer approaches the target layer, and that empirical measurements of the overall error also validate our theory. This provides an effective validation of our mathematical interpretation of the J-lens and explains the early failure of the J-lens from the perspective of mathematical bias: large nonlinear and distributional biases cause the mapped results at the target layer to deviate substantially from the confidence interval and lose their intended meaning.

%% file: 4_Methods.tex
\begin{figure}[t]
  \vspace{-5mm}
  \centering
  \includegraphics[width=1.0\textwidth]{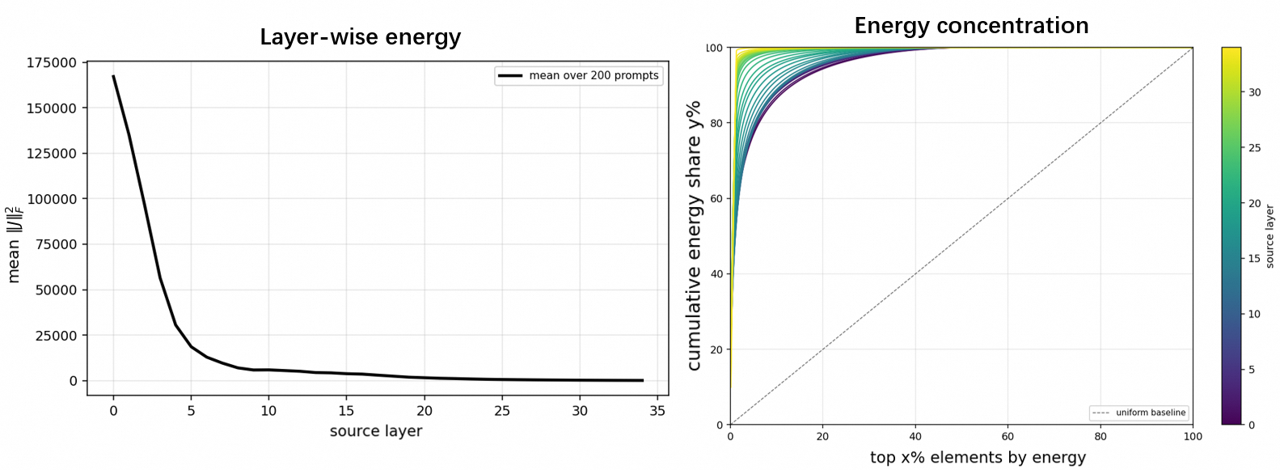}
  \caption{
  Jacobian energy analysis along the layers. The left figure indicates the relationship between absolute energy and the layers. The right figure indicates the energy concentration across layers, where the top x\% of elements accumulate y\% of the total energy of the layer.
  }
  \label{fig:layerwise-energy}
  \vspace{-2mm}
\end{figure}

\section{Jacobian Energy Structure in J-lens: Short Horizon and Sparse Concepts}\label{sec:energy}

This section moves from the validity of the averaged operator $J_\ell$ to the positional structure that the average hides. The theory above treats $J_\ell$ as an aggregate first-order readout, while we further ask which source-target pairs actually carry that aggregate.

\subsection{Setup}\label{subsec:energy-setup}
The positional Jacobian energy is adopted as a guide quantity, which measures how strongly a perturbation at one position propagates to a later final-layer position.
For layer $\ell$, source position $t$, and target position $t'$, define Jacobian energy as:
\begin{equation}
E_\ell(t,t')=\left\|\frac{\partial h_{\mathrm{final},t'}}{\partial h_{\ell,t}}\right\|^2.
\end{equation}
Since the J-lens is the average of the Jacobian matrices over all positions within a layer, analyzing the Jacobian energy at each position is important for understanding the representational tendency of the J-lens. It reveals which positions receive more attention from the J-lens, and the content at these positions with higher energy would influence the readout of the J-lens more strongly.

\subsection{Layer-wise decay and concentration}\label{subsec:decay-concentration}

As shown in Figure~\ref{fig:layerwise-energy}, two regularities across the layers appear before structural analysis. First, total energy decays with depth. Early layers sit upstream of many subsequent transformations and therefore have larger influences; late layers are closer to the output and act through shorter residual pathways. Second, energy is heavy-tailed across position pairs. A small top fraction of pairs accounts for most of the mass, and the concentration increases with depth. Thus the averaged object $J_\ell$ is not an even superposition of weak effects but the domination of a minority of position pairs. This also explains why the J-lens, although constructed from the output expectation over all future positions, can stably read out valid concepts.

\begin{figure}[t]
  \vspace{-5mm}
  \centering
  \includegraphics[width=1.0\textwidth]{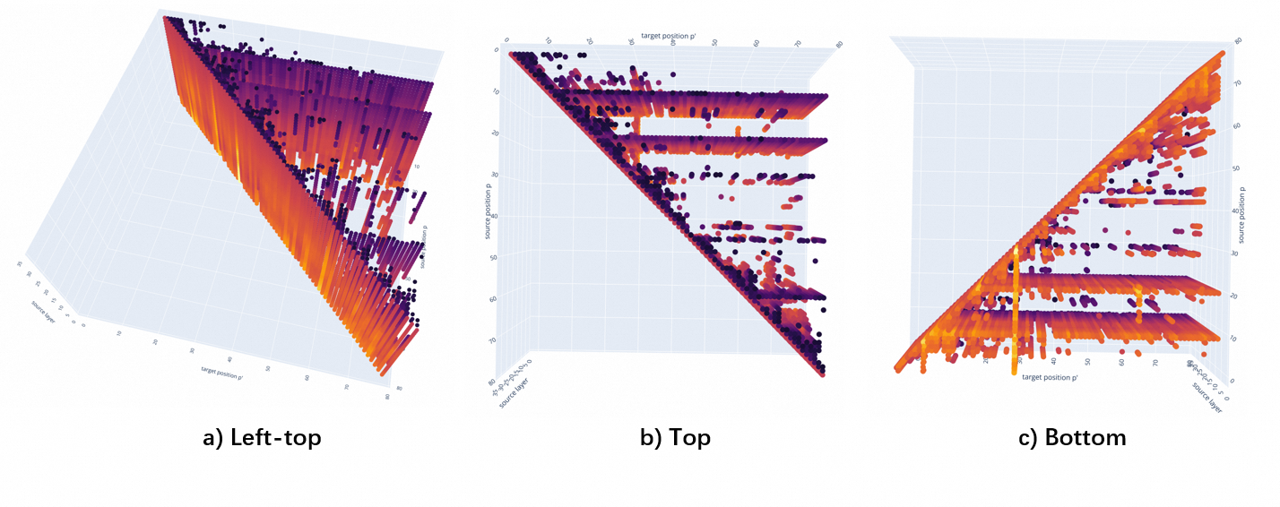}
  \caption{
  A 3D heatmap for the energy distribution of the Jacobian at each layer and position.
  }
  \label{fig:3d-heatmap}
  \vspace{-2mm}
\end{figure}

\subsection{Jacobian Structure: Short Horizons and Sparse Concept}\label{subsec:diag-concept}

To further investigate the distributional pattern of the highly concentrated Jacobian energy, we conduct visualization studies to analyze the structure of the J-lens and the causal implications of its actual representations. Specifically, we compute the Jacobian energy for every position pair in each layer during the construction of the J-lens and plot a three-dimensional heatmap over source positions, target positions, and layers. Since an overly dense point cloud is not suitable for display, and the concentration effect of energy distribution indicates that a small number of high-energy points are sufficient to reveal the structure, we display the 400 (approximately the top 13\%) position pairs with the highest energy in each layer, and hide the remaining points.

Figure~\ref{fig:3d-heatmap} demonstrates that high-energy pairs separate into two patterns. The first is the diagonal one: $E_\ell(t,t)$ is large, meaning that a position primarily preserves or prepares the next token. Diagonal energy is especially prominent in later layers, where the remaining computation is short and output-oriented. Since the elements on the diagonal and adjacent positions represent short-term expressions, these causal results are defined as short horizons. The second pattern is horizontal or vertical positions. Some source positions influence many future positions, behaving as broadcast origins; some target positions receive influence from many sources, behaving as integration points. When these positions are stable across prompts and attach to interpretable tokens, we group them as the sparse concept positions.

According to the visualization results above, let $t^*$ denote the positions of sparse concept, the formalization of J-lens can be approximated as:
\begin{equation}
J_\ell
=\mathbb{E}_{t,t'\ge t,\mathrm{prompt}}\left[\frac{\partial h_{\mathrm{final},t'}}{\partial h_{\ell,t}}\right]
\approx \mathbb{E}_{t,t'\approx t,\mathrm{prompt}}\left[\frac{\partial h_{\mathrm{final},t'}}{\partial h_{\ell,t}}\right]
+ 
\mathbb{E}_{t,t'= t^*,\mathrm{prompt}}\left[\frac{\partial h_{\mathrm{final},t'}}{\partial h_{\ell,t}}\right].
\end{equation}
Moreover, based on the derivation that interprets the J-lens as the aggregate expectation over future outputs in Section~\ref{sec:prelim}, we can reduce this future output to two positional modes: \textbf{short horizons} and \textbf{sparse concepts}. 
This is precisely the core understanding and interpretation of the meaning of the J-lens readout in this paper.

%% file: 5_Experiments.tex
\section{Experiments}

Based on the theoretical and experimental study above, we propose improvements to the J-lens to further validate our theory. Since experiments show that the Jacobian energy exhibits a pronounced concentration pattern, and that its concentration in the short horizon and sparse concept modes is precisely the design goal of the J-lens, the Jacobians at the remaining positions act as noise under this view. We therefore propose a simple but effective entry filtering strategy: only the Jacobian matrices at the top-j fraction of position pairs (p, p') with the largest energy are used to construct the J-lens. We also attempt to decouple the two readout modes of the J-lens, short horizon and sparse concept. Since heatmaps show that the Jacobian energy concentrates on the diagonal and on horizontal and vertical lines at specific positions, we apply masks with specific shapes to decouple the J-lens. For example, we apply a mask matrix that sets the diagonal to zero and keeps the remaining entries as one, with the expectation of retaining only sparse concepts. 

\subsection{Experiments Settings}
We utilized Qwen3-8B~\citep{qwen3} in the experiments. Following the previous study, Wikitext~\citep{wikitext} is introduced for J-lens construction, and six tasks including association, multihop reasoning, multilingual, ordering operations, poetry, and typos are evaluated~\citep{jlens}.

\subsection{Metrics}
\label{sec:metrics}
In this paper, we design two metrics named short-horizon layers (SHL) and intermediate-concept recall rate (ICR) to evaluate the capability of J-lens on next-token prediction and intermediate concept recalling individually.
Both metrics operate on the full position-layer lens readout: for a prompt
of $T$ tokens, the lens yields a vocabulary distribution
$\mathbf{y}^{(l)}_{t}$ at every layer $l \in \{1, \dots, L-1\}$ and position
$t \in \{1, \dots, T\}$.

\paragraph{Short-horizon layers.}
For each position $t$ with true next token $x_{t+1}$, we count the length of the trailing run of layers whose lens top-$1$ prediction matches $x_{t+1}$:
\begin{equation}
    SHL_t = \max \bigl\{ r : \arg\max_v q^{(l)}_{t,v} = x_{t+1}
  \ \forall\, l \in \{L - r + 1, \dots, L\} \bigr\}.
\end{equation}
Averaging $SHL_t$ over positions measures how early the lens readout converges to the model's next token prediction, demonstrating the short-horizon ability of a J-lens.

\paragraph{Intermediate-concept recall rate.}
For prompts annotated with intermediate concepts
$\mathcal{I} = \{w_1, \dots, w_m\}$, we count grid cells whose lens top-$K$ ($K = 10$) contains any token of concept:
\begin{equation}
  ICR = \bigl| \bigl\{ (l, t) : \mathcal{I} \cap
  \mathrm{top}\text{-}K\bigl(\mathbf{q}^{(l)}_{t}\bigr) \neq \varnothing
  \bigr\} \bigr|,
\end{equation}
which measures how broadly the J-lens reads the required sparse concept.

\begin{table*}[t]
  \vspace{-5mm}
\centering
\caption{Main results across several tasks. \textbf{Bold} denotes the best. Vanilla means the original J-lens. All J-lenses are trained on the same 200 examples with the same configuration.}
  \vspace{1mm}
\label{tab:main_results}
\resizebox{\linewidth}{!}
{
\begin{tabular}{lcccccccccccccc}
\toprule
 & \multicolumn{2}{c}{association} & \multicolumn{2}{c}{multihop} & \multicolumn{2}{c}{multilingual} & \multicolumn{2}{c}{order-ops} & \multicolumn{2}{c}{poetry} & \multicolumn{2}{c}{typo} & \multicolumn{2}{c}{avg} \\
\textbf{Method} & SHL & ICR & SHL & ICR & SHL & ICR & SHL & ICR & SHL & ICR & SHL & ICR & SHL & ICR\\
\midrule
\multicolumn{15}{c}{\emph{Qwen3-8B}} \\
\midrule
Vanilla & 1.01 & 4.70 & 1.65 & 16.58 & 0.66 & 40.07 & 0.66 & 12.82 & 0.99 & 19.54 & 0.62 & 14.40 & 0.94 & 18.71 \\
$\text{Filter}_{\text{topj}=0.5}$ & 1.04 & 4.93 & 1.71 & 15.58 & 0.68 & 38.53 & 0.68 & 12.56 & 1.00 & 20.17 & 0.67 & 15.69 & 0.98 & 18.60 \\
$\text{Filter}_{\text{topj}=0.2}$ & 1.05 & 6.36 & \textbf{1.73} & 16.35 & \textbf{0.69} & \textbf{48.40} & 0.67 & 17.36 & \textbf{1.06} & 21.13 & 0.69 & 19.03 & 1.00 & \textbf{22.14} \\
$\text{Filter}_{\text{topj}=0.1}$ & 1.06 & \textbf{6.44} & 1.72 & 15.78 & 0.68 & 47.38 & 0.68 & \textbf{18.26} & 1.04 & \textbf{21.17} & \textbf{0.72} & \textbf{19.25} & 1.00 & 22.00 \\
$\text{Filter}_{\text{topj}=0.01}$ & 1.05 & 4.28 & 1.69 & 13.26 & 0.67 & 31.56 & 0.69 & 11.82 & 1.03 & 18.96 & 0.69 & 15.73 & 0.98 & 16.45 \\
$\text{Filter}_{\text{w/o~Diag}}$  & 0.59 & 4.20 & 1.15 & \textbf{16.75} & 0.50 & 30.44 & 0.38 & 6.24 & 0.77 & 16.91 & 0.33 & 10.78 & 0.63 & 15.02 \\
$\text{Filter}_{\text{Diag}}$  & \textbf{1.08} & 4.72 & 1.70 & 13.28 & 0.68 & 33.61 & \textbf{0.74} & 12.75 & 1.05 & 18.84 & 0.69 & 16.11 & \textbf{1.00} & 17.07\\
\bottomrule
\end{tabular}
}
\end{table*}

\subsection{Results}
Results in Table~\ref{tab:main_results} presents the results for methods on several benchmarks in this paper. We can conclude the following finding. First, since the filtering methods with topj further strengthens the concentration of the Jacobian energy in the J-lens and enables its representational capacity to exceed that of the unfiltered baseline, J-lens can focus more on the highlight sparse concept and outperform the vanilla on most tasks.
Second, experiments on the diagonal filtering reveal that the coupling between the two patterns is deeper than the visualized results demonstrate: the ability of the J-lens to predict the next token is weakened but does not disappear, while its performance in generating sparse concepts also declines. This points to a direction for future work.

%% file: 6_Conclusion.tex
\section{Conclusion}\label{sec:conclusion}
We interpret the J-lens as an averaged first-order transfer operator from intermediate activations to expected future readouts: the Jacobian matrix is the locally optimal linearization, least squares gives the globally optimal affine readout, and the Stein bridge makes them coincide under Gaussian inputs. Nonlinear curvature and non-Gaussian score residuals explain systematic bias and provide an explanation for the early failure of the J-lens. We then further analyze the structure of the Jacobian energy. The Jacobian energy decays with depth, concentrates in a small number of source-target pairs, and splits into diagonal short horizons and sparse concept positions. Finally, we introduce improvements and decoupling methods for the J-lens. Energy filtering preserves or improves next-token consistency and the recall rate of intermediate concepts, thereby validating our interpretation of the meaning of the J-lens readout. 
Finally, we acknowledge that this work still has limitations and is still in progress, which does not represent the final results.